\documentclass[a4paper,fleqn]{cas-sc}

\usepackage[authoryear,longnamesfirst]{natbib}

\def\tsc#1{\csdef{#1}{\textsc{\lowercase{#1}}\xspace}}
\tsc{WGM}
\tsc{QE}
\tsc{EP}
\tsc{PMS}
\tsc{BEC}
\tsc{DE}

\begin{document}
\let\WriteBookmarks\relax
\def\floatpagepagefraction{1}
\def\textpagefraction{.001}
\shorttitle{Counterfactual Samples Constructing for Commonsense Statements Estimation}
\shortauthors{C Liu et~al.}

\title [mode = title]{Counterfactual Samples Constructing and Training for Commonsense Statements Estimation}

\author[1]{Chong Liu}[style=chinese] 

\fnmark[1]

\credit{Writing – original draft, Conceptualization, Methodology, Investigation, Software}

\author[1]{Zaiwen Feng}[style=chinese] 
\fnmark[1]

\credit{Writing – review \& editing, Conceptualization, Methodology}

\author[2]{Lin Liu}[style=chinese] 

\credit{Writing – review \& editing, Conceptualization}
\author[3]{Zhenyun Deng}[style=chinese] 

\credit{Writing – review \& editing, Methodology}

\author[2]{Jiuyong Li}[style=chinese]

\credit{Writing – review \& editing, Methodology}

\author[1]{Ruifang Zhai}[style=chinese] 

\credit{Writing – review \& editing, Supervision, Validation}

\author[2]{Debo Cheng}[style=chinese] 
\cormark[1]
\ead{debo.cheng@unisa.edu.au}
\credit{Writing – review \& editing, methodology, Project administration, Conceptualization}

\author[1,4]{Li Qin}[style=chinese] 
\cormark[1]
\ead{qinli@mail.hzau.edu.cn}
\credit{Writing – review \& editing, Conceptualization, Supervision, Funding acquisition}

\affiliation[1]{organization={College of Informatics },
            addressline={Huazhong Agricultural University}, 
            city={Wuhan},
            postcode={430070}, 
            state={Hubei},
            country={China}}
\affiliation[2]{
   UniSA STEM, 
    University of South Australia, 
   Adelaide, 5095, South Australia, Australia
}
\affiliation[3]{
   Department of Computer Science and Technology, 
    University of Cambridge, 
    Cambridge, CB2 1TN, United Kingdom
}
\affiliation[4]{
    organization={Key Laboratory of Smart Farming for Agricultural Animals},
    addressline={Huazhong Agricultural University}, 
    city={Wuhan},
    postcode={430070}, 
    state={Hubei},
    country={China}
}

\fntext[fn1]{Contributed equally to this work.}
\cortext[cor1]{Corresponding author}

\begin{abstract}
Plausibility Estimation (PE) plays a crucial role for enabling language models to objectively comprehend the real world. While large language models  (LLMs) demonstrate remarkable capabilities in PE tasks but sometimes produce trivial commonsense errors due to the complexity of commonsense knowledge. They lack two key traits of an ideal PE model: a) \textit{Language-explainable}:  relying on critical word segments for decisions, and b) \textit{Commonsense-sensitive}: detecting subtle linguistic variations in commonsense. To address these issues, we propose a novel model-agnostic method, referred to as  \textbf{C}ommonsense \textbf{C}ounterfactual \textbf{S}amples \textbf{G}enerating (\textbf{CCSG}). By training PE models with CCSG, we encourage them to focus on critical words, thereby enhancing both their language-explainable and commonsense-sensitive capabilities. Specifically, CCSG generates counterfactual samples by strategically replacing key words and introducing low-level dropout within sentences. These counterfactual samples are then incorporated into a sentence-level contrastive training framework to further enhance the model’s learning process. Experimental results across nine diverse datasets demonstrate the effectiveness of CCSG in addressing commonsense reasoning challenges, with our CCSG method showing 3.07$\%$ improvement against the SOTA methods.
\end{abstract}

\begin{keywords}
    Plausibility Estimation \sep Counterfactual Reasoning \sep Large Language Models \sep Commonsense Biases \sep Contrastive Learning
\end{keywords}

\maketitle

\section{Introduction}
Plausibility Estimation (PE), which assesses the plausibility of natural language sentences according to general knowledge that people commonly possess, is one of the fundamental capabilities of advanced Artificial intelligence (AI) agents. With the advancement of large language models (LLMs), PE has made considerable progress in distinguishing between commonsense and non-commonsense statements. However, due to inevitable manual annotation artifacts in real world datasets, existing PE models often fall into over-reliance on superficial linguistic correlations between sentences and their associated golden labels (also known as language biases) \citep{bender2020climbing, marcus_davis_2023, talmor2018commonsenseqa}. For example, a model might childishly answer ``\emph{false}'' for ``\emph{Mr. July ordered wires for dinner at a Chinese restaurant}'' and still receive a satisfactory estimation.

\begin{figure}
    \centering
    \includegraphics[width=0.5\linewidth]{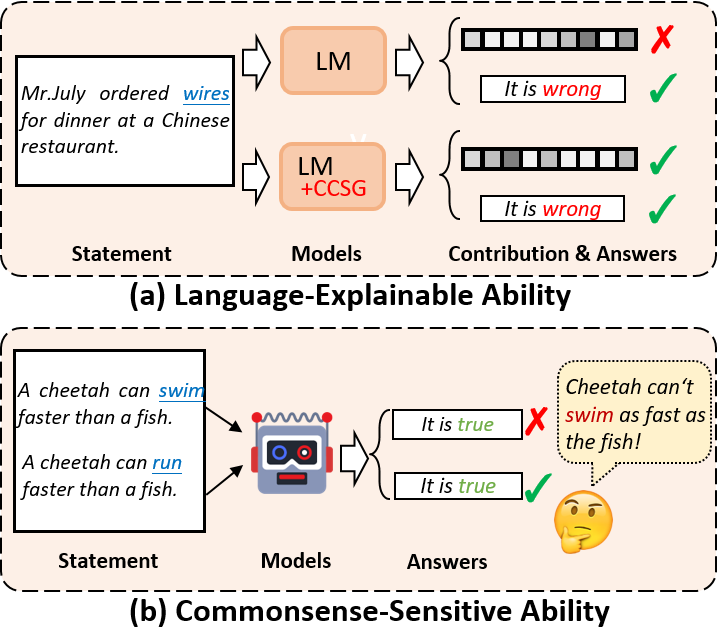}
    \caption{An ideal PE model should exhibit two indispensable characteristics: (a) \textbf{Language-Explainable ability}: The PE model should not only make a correct prediction but also base its prediction on the appropriate linguistic reference regions. (b) \textbf{Commonsense-Sensitive ability}: The PE model must be enough sensitive to commonsense variations in texts. For example, replacing the critical word ``swim'' with ``run'', in a statement should result in opposite predicted labels, reflecting the underlying change in plausibility.}
    \label{fig:fig01}
\end{figure}

Currently, the most prevalent solutions for PE fall into two categories: a) \emph{Relying on external knowledge} \citep{thorne2018fever, ling2023knowledge}: These approaches leverage neural language models to extract reasoning chains from external knowledge bases, pinpointing solutions to commonsense questions while providing supporting knowledge assertions. However, they are heavily dependent on the availability and quality of external knowledge bases, which are costly and time-intensive to construct.
b) \emph{Construction-based methods} \citep{zeng2020counterfactual, wang2021robustness, udomcharoenchaikit2022mitigating, tokpo2024fairflow}: These methods automatically generate parallel counterfactual data by intervening in existing observational examples to enhance the original dataset. The primary aim is to focus the model’s attention on critical samples, thereby strengthening the dataset’s quality and informativeness.

Although construction-based methods show remarkable performance, they neglect to endow models with the two crucial characteristics necessary for an ideal PE mode: a) \textbf{Language-Explainable ability}: An ideal PE model should depend on the correct linguistic reference segments, adhering to the principle of being \textit{right for the right reasons} \citep{zeng2020counterfactual, ross2017right}. As shown in Figure~\ref{fig:fig01}(a), while two models might both predict ``\emph{It is wrong}'' correctly, they may rely on different linguistic regions, leading to differing levels of explainability. b) \textbf{Commonsense-Sensitive ability}: An ideal PE model must be enough sensitive to commonsense linguistic variations in statements. For instance, as shown in Figure~\ref{fig:fig01}(b), similar statements with identical structures (where the word ``\emph{swim}'' is replaced by ``\emph{run}''), should yield different predictions, reflecting the change in background commonsense. The model must recognize this discrepancy and adjust its predictions accordingly.

In this research, we propose a novel method, \textbf{C}ommonsense \textbf{C}ounterfactual \textbf{S}amples \textbf{G}enerating (\textbf{CCSG}), which could be used for a plug-and-play component to enhance the language-explainable and commonsense-sensitive abilities of PE models. Specifically, we develop a counterfactual samples constructor using low-level dropout and word-piece replacement, guided by word contributions. Our CCSG method is independent of pre-trained knowledge, making it a versatile mediator to guide language models (LMs) toward focusing on relevant linguistic regions while improving commonsense sensitivity. Additionally, CCSG significantly enhances the estimation performance of LMs on PE tasks.

In summary, this paper makes the following key contributions:
\begin{itemize}
\item  We introduce a contrastive loss based on counterfactual samples and design a novel mechanism for generating positive and negative commonsense counterfactual samples.
\item We leverage the Structural Causal Model (SCM) to analyze the commonsense bias from a causal perspective and propose using a counterfactual samples constructor as an intermediary to mitigate this bias through causal inference techniques.
\item  The proposed CCSG method has been validated on 9 distinct datasets, demonstrating its effectiveness in reducing bias and achieving new benchmarks in performance for the PE task.
\end{itemize}

\section{Related Work}
In this section, we review key advancements and methodologies in PE, causality for language processing, and spurious correlation, highlighting their relevance to our work.

\subsection{Plausibility Estimation}
Previous studies have extensively explored the concept of verifying commonsense statements. Models such as I2D2 \citep{bhagavatula2022i2d2} were designed to evaluate the validity of commonsense statements generated by other models. Similarly, the ENTAILER model \citep{tafjord2022entailer} assesses the validity of provided hypotheses, albeit with incomplete training, limiting its robustness. However, these models, trained on limited and domain-specific datasets, show restricted applicability across broader commonsense domains. Additionally, some efforts \citep{liu2023vera} have employed pretrained language models (PLMs) to validate commonsense sentences \citep{jung2022maieutic}.

\subsection{Causality for Language Processing}
Recent studies have increasingly explored the integration of causal inference with language models in the field of NLP \citep{hu2021causal}. Applications include tasks like controllable text generation \citep{madaan2021generate, goyal2020cam} and counterfactual reasoning \citep{mu2023enhancing, chen2023causal}. Moreover, causal inference techniques, particularly Structural Causal Models (SCMs) \citep{pearl2009causality,cheng2024data}, have been employed to identify spurious correlations and address language bias through causal intervention methods \citep{feng2023less, wang2023causal, zevcevic2023causal}. Compared to traditional approaches, causal inference has demonstrated significant advantages in debiasing within NLP applications.

\subsection{Spurious Correlation}
Spurious correlations are a significant challenge and can arise in various ways \citep{kiritchenko2018examining, roemmele2011choice, keith2020text, DBLP:journals/ipm/WangSMXWYS25}. An increasing body of research is devoted to understanding the complexities of spurious correlations and leveraging causal inference strategies to enhance model robustness in deep learning. For instance, \citet{wood2018challenges} utilized text classification techniques within causal analysis frameworks to tackle challenges like data loss and measurement inaccuracies. Similarly, \citet{roberts2020adjusting} proposed methods for mitigating confounding effects in causal estimates. To enhance model robustness against adversarial perturbations, \citet{jia2019certified} introduced label-preserving transformations using Interval Bound Propagation.
In addition, \citet{wang2020identifying} developed classifiers for the purpose of differentiating between spurious and authentic features, progressively weakening spurious features to enhance worst-case evaluations for minority groups. \citet{eisenstein2022informativeness} utilized domain knowledge to identify spurious correlations that pose significant threats to model robustness. Finally, \citet{ribeiro2020beyond} assessed model robustness through the generation of counterfactual examples, which require expert human intervention.

\section{Preliminaries}

\subsection{Task Definition}
In this study, we formulate the problem of plausibility estimation as a binary classification challenge. Given a commonsense statement $x$, the proposed model generates a real-valued confidence score $s$, within the range $[0, 1]$. This score represents the model’s estimation of the plausibility of $x$. Although the ground-truth label is binary, the PE model produces a continuous score to express its level of certainty. A score of 1.0 signifies the model’s absolute conviction in the accuracy of $x$, while a score of 0.0 represents complete certainty of its incorrectness. To derive a binary prediction from the confidence score, a threshold of 0.5 is used.

\subsection{Causal Graph Model}
Identifying which segments of a sentence represent commonsense information, as well as classifying these segments as background knowledge support, is a challenging task. These are inherently causal inquiries, requiring an understanding of the data generation process that extends beyond mere observational data \citep{pearl2009causal}. Selection biases present in observational data frequently result in spurious correlations, which hinder the generalization capabilities of PE models, particularly in scenarios with limited data. To address this challenge, causal analysis provides a nuanced framework that uncovers the underlying mechanisms and mitigates such biases effectively.

To rigorously analyze the causal dynamics between the PE model and the underlying data, we employ causal directed acyclic graphs (DAGs) \citep{pearl2009causal,cheng2024data} to represent the data generation process. Building on the proposed causal DAG, we utilize a Structural Causal Model (SCM) \citep{pearl2000models} to delineate the inferential mechanisms of the PE model. In the causal DAG, nodes correspond to random variables, while directed arcs indicate causal influence from one variable to another~\citep{pearl2000models}.

\begin{figure}
    \centering
    \includegraphics[width=0.67\linewidth]{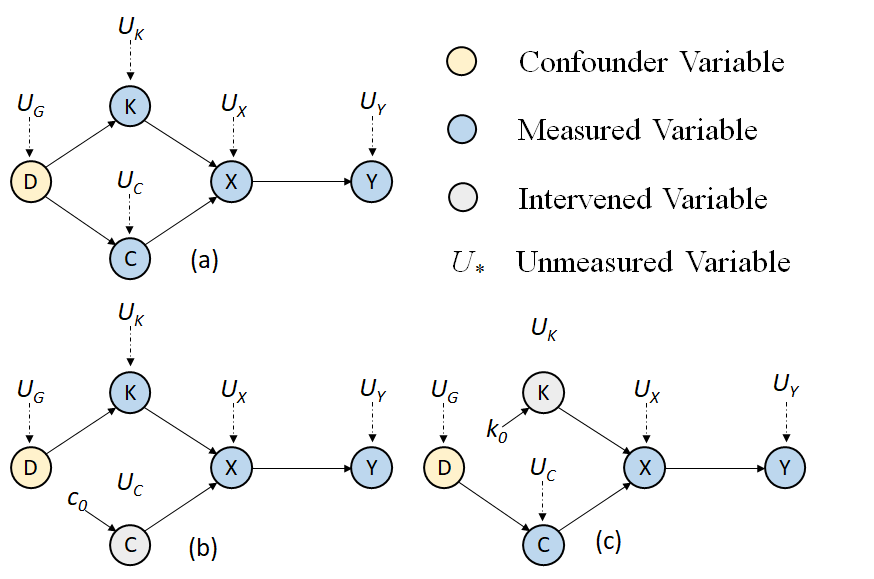}
    \caption{The SCM depicts the inferential mechanism of the PE model without any interventions. (b) An intervention on the variable $C$ is assigned the value $c_0$, represented as do($C = c_0$). Similarly, an intervention on the variable $K$ is assigned the value $k_0$, represented as do($K = k_0$)}
    \label{fig:fig02}
\end{figure}

For analytical clarity, we decompose the sentence into two constituent variables: the keyword ($K$) and the context ($C$). The proposed causal DAG is illustrated in Figure~\ref{fig:fig02}(a). Using this SCM framework, we construct our CCSG method as follows:
\begin{equation}
    \begin{split}
        d&:=f_D(U_D)\\
        k&:=f_K(d,U_K)\\
        c&:=f_C(d,U_C)\\
        x&:=f_X(k,c,U_X)\\
        y&:=f_Y(x,U_X)
    \end{split}
\end{equation}
where $D$ denotes a confounding variable that exerts influence on the generation of both the keyword $K$ and the contextual commonsense information $C$. The input example, denoted as $X$, is produced as a result of the interaction between $K$ and $C$. Subsequently, $Y$ represents the evaluation metric, specifically the accuracy score, which quantifies the performance of the PE model. Additionally, $U^*$ signifies the presence of unmeasured variables that may impact our observations.

Causal effects are useful tool in elucidating the causal relationships that underpin the dynamics of a system~\citep{cheng2023causal}. The estimation of these causal effects is fundamentally rooted in the simulation of interventions within the SCM. To this end, we employ the mathematical operator $do(v_0)$~\citep{pearl2009causal}, which allows us to mimic physical interventions by setting the variable $v$ to a specific value, $v_0$. For example, to conduct an intervention $do(c_0)$ within the SCM denoted by $M$, we stabilize the variable $C$ at the value $c_0$, as illustrated in~\ref{fig:fig02}(b). This intervention is mathematically represented as:
\begin{equation}
    C:=c_0
\end{equation}

This intervention effectively severs the causal influence of the confounding variable $D$ on the variable $C$. Consequently, the post-intervention distribution, denoted as $P(y|do(C = c_0))$, provides an estimation of the proportion of individuals that would exhibit a response at level $Y = y$ under the hypothetical scenario where the treatment $C$ is uniformly set to $c_0$ across the entire population \citep{pearl2009causal}. In the context of our specific analysis, we find that $P(y|do(C = c_0))$ indicates a uniform response level within the population under the specified intervention.

An approach to estimate the treatment effect or causal influence is to measure the mean difference between the distributions before and after the intervention. This is achieved by employing the expectation operator $\mathbf{K}$, which gives rise to the concept of the Average Causal Effect (ACE). The ACE is mathematically represented as:
\begin{equation}
    ACE_C = \textbf{K}(y|do(c_0)) - \textbf{K}(y|do(c))
\end{equation}
where $c_0$ represents the intervened value of the treatment $C$, and $c$ represents its original, unaltered value. Similarly, to estimate the effects of the variable $K$ on the variable $Y$, we can perform an intervention on the variable $K$, denoted as $do(k_0)$ (see Figure~\ref{fig:fig02}(c)).

\section{The Proposed CCSG Method}

Our CCSG method automatically substitutes entities within observational instances to generate novel counterfactual examples. This approach addresses the limitations of language-explainable and commonsense-sensitive features inherent in limited observational data, thereby enhancing the PE model’s ability to identify more invariant and robust features.

\begin{figure*}
    \centering
    \includegraphics[width=0.95\textwidth]{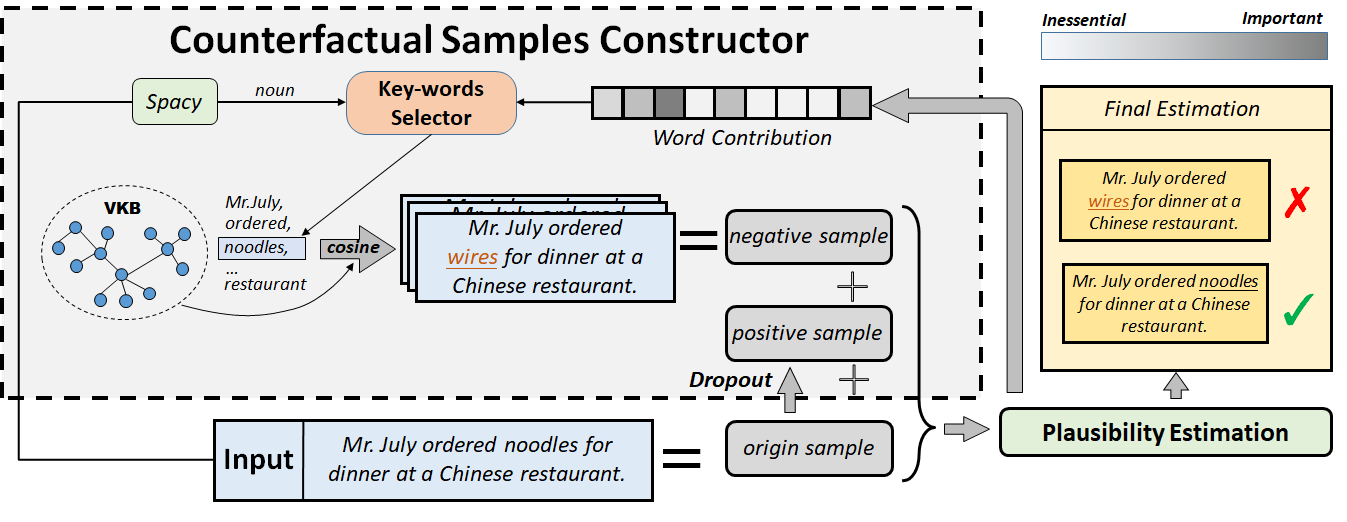}
    \caption{The overview of our CCSG framework is as follows: 1) We prepare appropriate keywords based on the word contributions from the previous training output. 2) We replace the keywords using a vector knowledge base (VKB), such as \emph{Glove}, to generate counterfactual negative samples, and generate counterfactual positive samples by applying dropout \citep{gao2021simcse} to the original samples. 3) We send the counterfactual samples to the PE model, where sentence-level contrastive learning is used to enhance the model’s commonsense reasoning ability.}
    \label{fig:fig03}
\end{figure*}

The overall framework of CCSG is illustrated in Figure \ref{fig:fig03} and comprises two major modules: 1)	Counterfactual Samples Constructor: Generates counterfactual samples by replacing words based on their contribution, and 
	2)	Plausibility Estimation: Trains a T5-5b-only-encoder extractor using the constructed counterfactual samples to estimate commonsense statements from input text. In the following sections, we introduce the two modules of CCSG in detail.

\subsection{Counterfactual Samples Constructor}
The fundamental concept underpinning our method is the identification of an alternative keyword to intervene on a given word within an observational sample. However, compiling a keyword set tailored to a particular domain requires substantial human effort to curate words that exhibit no discernible difference.

\begin{enumerate}
  \item \textbf{Initial Entities Selection.} Generally, for any particular statement, only a few entities in statement $x$ are relevant. To refine the criteria for identifying critical entities, we initially curate a more compact set of candidates, denoted as $I$. In the absence of manually annotated data delineating the pivotal entities for individual samples, we utilize the approach detailed by \citet{chen2021explaining} to extract entities with a high affinity to the PE model. Explicitly, we first applies the spaCy tagger \citep{honnibal2017spacy} to label each word in the statement with its corresponding Part-of-Speech (POS) tag, followed by the extraction of nouns from the text. Subsequently, The cosine similarity between the GloVe \citep{pennington2014glove} embeddings of the entities and the identified nouns is computed, resulting in a set of similarity scores, termed $sim$, for all words within $I$.
  Ultimately, we distill the initial word set $I$ by picking the top $|I|$ words with the largest $sim$ scores.
  \item \textbf{Word Contributions Calculation.} 
  After the acquisition of the initial set of words $I$, the contribution of each word on the predicted score of the correct label is computed.
  Based on recent studies \citep{selvaraju2019taking, jain2019attention, DBLP:journals/ipm/ZhaoXJT24}, using modified Grad-CAM \citep{selvaraju2020grad} to identify the impact of each word. We compute the 
  word contribution of the $i$-th word embedding to the correct answer $\alpha$ as follows:
\begin{align}
s(\alpha, e_i) = S(P_{pe}(\alpha), e_i) = \left( \frac{\partial P_{pe}(\alpha)}{\partial e_i} \right)^T  \times \mathbf{1}
\end{align}
where $e_i$ is the $i$-th word embedding, $P_{pe}(\alpha)$ is the predicted probability of the correct label answer $\alpha$, and $\textbf{1}$ is a vector to obtain the sum of partial derivatives. Moreover, if the score $s(\alpha, e_i)$ is higher, the contribution of word $e_i$ to 
its correct answer $\alpha$ is larger.
  \item \textbf{Critical Entities Substitution.} After calculating the contribution scores $s(\alpha, e_i)$ for each word in $I$, we pick out the top-K words with the largest scores to serve as keywords for substitution and masking. Specifically,we find words in the VKB (e.g., GloVe, FastText) that have high cosine similarity with the keyword as alternative words and replace them. The negative samples produced at this stage comprise counterfactual instances imbued with anti-commonsense data, thereby enhancing the model’s acuity in detecting and responding to commonsense information.

  \item \textbf{Positive Samples Generation.} Generating accurate positive samples and ensuring high-quality output is challenging, and difficult to be achieved through substitution or masking. Therefore, we introduce the application of low-level dropout \citep{gao2021simcse}, an effective method for generating positive samples.
\end{enumerate}

\subsection{Plausibility Estimation}
\subsubsection{Model Architecture} In our approach, we employ a Transformer-based LLM as the core of CCSG, designed to generate a real-valued score $s$ within the range of $[0, 1]$ for a given statement $x$. To obtain the input representation, we leverage the last hidden state $h$ associated with the EOS token. The choice of EOS is based on its ability to encapsulate the entire input in bidirectional encoder architectures, such as Flan-T5 encoder. After that, a MLP layer projects the vector $h$ onto a scalar logit $z$, which is then processed by a sigmoid function $\sigma(\cdot)$, converting it into a predictive score $s$. The process can be formally expressed as follows:
\begin{equation} 
h=f_{LM}(x),~z=f_{MLP}(h),~s=\sigma(z). 
\end{equation}

\noindent 
where $h(x)$ represents the embedding of the input statement $x$. $z(x)$ denotes the logit of $x$, and $s(x)$ denotes the prediction score.

\subsubsection{Batch Setting} The batch is comprised of statements from various statement groups, with all statements from the same group being grouped together within the batch. The $B_G$ denotes the count of statement groups and the $B_S$ denotes the total count of statements in a batch. The statement groups are denoted as$\{X_j\} ^{B_G}_{j=1}$, and the statements as $\{x_i\}^{B_S}_{i=1}$ is a partition of $\{X_i\}^{B_S}_{j=1}$. The correctness label of each statement $x_i$ is  $y_i\in\{0,1\}$.

\subsubsection{Loss Calculation} The training loss of CCSG consists of two parts: a binary classification loss and a supervised contrastive loss, expressed as $\mathcal{L} = \alpha \mathcal{L}_{bin} + \beta \mathcal{L}_{cot}$.

\vspace{2mm}

\noindent \textbf{Binary Classification Loss.}
 Validating commonsense statements could be considered as a task of binary classification problem. The objective is to minimize the binary classification loss function, computed as follow:
\begin{align} 
\mathcal{L}_{bin} & =-a_i\log P_{PE}(e_i)-(1-a_i)\log(1-P_{PE}(e_i)), 
\end{align}

\noindent where $a_i$ denotes the accurate classification for the $i$-th sample, $e_i$ represents the feature associated with the $i$-th word, and $P_{PE}$ signifies the predicted probability of the $i$-th sample's correct label.

\vspace{2mm}

\noindent \textbf{Supervised contrastive loss.~} 
\citet{khosla2020supervised} have shown that employing supervised contrastive learning methods bolsters a model's resilience and ability to generalize when confronted with diverse input data.
Consequently, we have incorporated supervised sentence-level contrastive learning into our methodology. This strategic enhancement is applied to the input representations $h$. For each golden statement $x_i$ within a batch, the objective of the contrastive loss is to maximize the similarity between $x_i$ and every counterfactual positive sample $x_p$ that shares the same correctness label as $x_i$, treating them as positive examples. Concurrently, the loss seeks to minimize the similarity between $x_i$ and all counterfactual negative statements $x_n$ that possess an opposing correctness label to $x_i$, considering them as negative examples. The formulation of the supervised contrastive loss \citep{liu2023vera} is as follows:
\begin{equation}
    \begin{aligned} &\mathcal{L}{cot}= -\log \frac{\sum{k \in \mathcal{P}(i)} \exp [\cos (\boldsymbol{h}\left(x_{i}\right), \boldsymbol{h}\left(x_{k}\right)) / \tau]}{\sum_{k \in \mathcal{P}(i) \cup \mathcal{N}(i)} \exp [\cos (\boldsymbol{h}\left(x_{i}\right), \boldsymbol{h}\left(x_{k}\right)) / \tau]}, \end{aligned}
\end{equation}

\noindent where $\tau$ denotes the contrastive learning temperature hyperparameter, $\cos{(\cdot,\cdot)}$ signifies the cosine similarity function, $\mathcal{P}(i)\subseteq [B_{S}]$ is the index set of counterfactual positive examples for $x_i$, and $\mathcal{N}(i)\subseteq [B_{S}]$ is the index set of counterfactual negative examples for $x_i$.
\begin{equation}
    \begin{aligned}
        \mathcal{P}(i) &= \{m \mid 1 \leq m \leq B_S, y_m = y_i, m \neq i\}, \\
        \mathcal{N}(i) &= \{m \mid 1 \leq m \leq B_S, y_m \neq y_i\}.
    \end{aligned}
\end{equation}

\section{Experiment Settings}
This section delineates the specifics of model training, the evaluation methodologies and metrics applied, as well as the baseline models that were utilized for comparative performance analysis.

\begin{table*}[!ht]
    \centering
    \footnotesize
    \renewcommand{\arraystretch}{1.1}
    \setlength{\tabcolsep}{1mm}{
    \caption{\label{tab:tab05}
         Datasets and statistics. 
         The figure enclosed in parentheses beneath the Format heading indicates the quantity of options per data.
         The final three columns display the count of total, true and False statements in the test distribution.
         Table \ref{tab:tab06} shows statement length statistics, and Table~\ref{tab:tab07} displays full citations and links for all datasets.
    }
    \begin{tabular}{lcccrrrrr}
    \toprule
        \textbf{Abbreviation} & \textbf{Full Name} & \textbf{Domain} & \textbf{Format} & \textbf{Train} & \textbf{Test} & \textbf{Statements} & \textbf{S-True} & \textbf{S-False} \\ \hline
        OBQA & OpenBookQA & scientific & multi-option(4) & 4,950 & 500 & 2,000 & 500 & 1,500 \\
        COPA & COPA & commonse & multi-option(2) & 400 & 100 & 200 & 50 & 150 \\
        SciQ & Scientific Question & scientific & multi-option(4) & 11,620 & 1,000 & 4,000 & 1,000 & 3,000 \\
        QASC & QASC & scientific & multi-option(8) & 8,130 & 930 & 7,444 & 930 & 6,514 \\
        SIQA & Social-IQA & social & multi-option(3) & 33,140 & 1,950 & 5,860 & 1,955 & 3,905 \\
        CODAH & CODAH & commonse & multi-option(4) & 2,200 & 576 & 2,304 & 576 & 1,728 \\
        ComVE & ComVE (Task A) & commonse & multi-option(2) & 9,997 & 1,000 & 2,000 & 1,000 & 1,000 \\
        CSQA & CommonsenseQA-1.0 & commonse & multi-option(5) & 9,741 & 1,221 & 6,099 & 1,221 & 4,878 \\multi-option
        CSQA2 & CommonsenseQA-2.0 & commonse & bool & 9,264 & 2,541 & 2,541 & 1,225 & 1,316 \\
        C2S & Com2Sense-paired & commonse & multi-option(2) & 805 & 390 & 780 & 390 & 390 \\
        PIQA & Physical-IQA & physical & multi-option(2) & 16,111 & 1,840 & 3,640 & 1,840 & 1,840 \\
        WG & Winogrande & commonse & multi-option(2) & 40,395 & 1,270 & 2,540 & 1,270 & 1,270 \\
        \textbf{Total} & - & - & - & \textbf{147,100} & \textbf{13,311} & \textbf{39,399} & \textbf{11,945} &  \textbf{27454} \\ \bottomrule
    \end{tabular}}
\end{table*}

\begin{table*}[!h]
    \centering
    \footnotesize
    \renewcommand{\arraystretch}{1.1}
    \setlength{\tabcolsep}{1.2mm}{
    \caption{\label{tab:tab06}
        Length distribution statistics of all datasets. This table presents the distribution of statement lengths across the datasets.
    }
    \begin{tabular}{llrrrrrr}
    \toprule
        \textbf{Abbreviation} & \textbf{Name} & \multicolumn{6}{c}{\textbf{Length Distribution}} \\
        & & min & median & 90$\%$ & 95$\%$ & 99$\%$ & max \\ \hline
        OBQA & OpenBookQA & 5 & 16 & 29 & 36 & 56 & 74 \\
        COPA & COPA & 10 & 17 & 21 & 23 & 26 & 28 \\
        SciQ & SciQ  & 6 & 19 & 29 & 34 & 48 & 75 \\
        QASC & QASC  & 5 & 13 & 19 & 21 & 24 & 30 \\
        SIQA & Social IQA  & 10 & 28 & 38 & 41 & 51 & 70 \\
        CODAH & CODAH  & 5 & 21 & 31 & 34 & 45 & 73 \\
        ComVE & ComVE (Task A) & 4 & 10 & 14 & 16 & 20 & 28 \\
        CSQA & CommonsenseQA & 5 & 18 & 28 & 32 & 43 & 73 \\
        CSQA2 & CommonsenseQA 2.0 & 5 & 14 & 24 & 29 & 38 & 58 \\
        C2S & Com2Sense (paired)  & 12 & 24 & 34 & 38 & 44 & 55 \\
        PIQA & Physical IQA  & 5 & 26 & 62 & 80 & 120 & 256 \\
        WG & Winogrande & 17 & 24 & 31 & 34 & 38 & 42 \\ \bottomrule
         
    \end{tabular}}
\end{table*}

\begin{table*}[!h]
    \centering
    \footnotesize
    \renewcommand{\arraystretch}{1.1}
    \setlength{\tabcolsep}{1.2mm}{
    \caption{\label{tab:tab07}
        More dataset source details. We provide the link for each dataset we sourced, and indicate whether training data for Flan-T5 includes these datasets.
    }
    \begin{tabular}{llllc}
    \toprule
        \textbf{Abbr.} & \textbf{Name} & \textbf{Citation} & \textbf{Link} & \textbf{In Flan-T5?} \\ \hline
        OBQA & OpenBookQA &  \url{https://github.com/allenai/unifiedqa} & yes \\
        COPA & COPA &  \url{https://huggingface.co/datasets/super_glue} & yes \\ 
        SciQ & SciQ  &  \url{https://allenai.org/data/sciq} & yes\\
        QASC & QASC  & \url{https://github.com/allenai/qasc} & yes \\
        SIQA & Social IQA &  \url{https://github.com/allenai/} & yes \\
        CODAH & CODAH  &  \url{https://github.com/Websail-NU/CODAH} & yes \\
        ComVE & ComVE (Task A)  & \url{https://github.com/wangcunxiang} & no \\
        CSQA & CommonsenseQA &  \url{https://github.com/allenai/csqa} & yes \\
        CSQA2 & CommonsenseQA 2.0 &  \url{{https://github.com/allenai/csqa2}} & yes \\
        C2S & Com2Sense (paired) &  \url{https://github.com/PlusLabNLP/Com2Sense} & yes \\
        PIQA & Physical IQA &  \url{https://github.com/allenai} & yes \\
        WG & Winogrande  &\url{https://github.com/allenai/winogrande} & yes \\ \bottomrule

    \end{tabular}}
\end{table*}

\begin{table*}[!ht]
    \centering  
    \footnotesize
    \renewcommand{\arraystretch}{1.2}
    \setlength{\tabcolsep}{1.2mm}{
    \caption{\label{tab:bert}
        Prediction results of BERT on PE. The prediction scores are close to the proportion of ``True'' and ``False'' labels in the dataset which shows that the traditional BERT is difficult to learn commonsense features in PE tasks.
    }
    \begin{tabular}{lcccccccccc}
    \toprule
        \textbf{Dataset} & \textbf{Avg} & OBQA & COPA & SciQ & QASC & SIQA & COHDA & ComVE & CSQA2 & C2S\\
                  &   & 1:3 & 1:3 & 1:3 & 1:7 & 1:2 & 1:3 & 1:1 & 1:1 & 1:1\\ \hline
        BERT & 34.79 & 25.25 & 31.10 & 32.55 & 13.75 & 33.91 & 26.12 & 51.02 & 49.22 & 50.17   \\ \hline
        ~~~~+~LSTM & 35.35 & 25.82 & 32.06 & 32.21 & 13.78 & 34.54 & 28.80 & 51.24 & 48.18 & 51.52\\
        ~~~~+~CNN & 36.95 & 27.99 & 34.51  & 33.15 & 12.63 & 34.86 & 29.51 & 54.34 & 53.67 & 51.89\\ \bottomrule
    \end{tabular}}
\end{table*}

\subsection{Datasets}
We rigorously evaluated the efficacy of our approach using nine distinct datasets specifically designed for PE tasks. Furthermore, we extended our evaluation to include its performance in LLM commonsense filtering, utilizing an additional six datasets, as detailed in Table~\ref{tab:tab05}.

Table~\ref{tab:tab06} presents the distribution of statement lengths across all datasets, providing insights into the diversity of input text lengths. Additionally, Table~\ref{tab:tab07} includes citations and links to the datasets, ensuring transparency and facilitating reproducibility of the experiments.

\subsection{Base Models}
We utilize the encoder module of Flan-T5 \citep{raffel2020exploring} as the backbone for CCSG. Specifically, we start with the pre-trained T5-v1.1-XXL variant, which features an encoder with approximately 5 billion parameters. Additionally, we conducted experiments using pretrained BERT models (small and medium variants) and found that they struggled with PE tasks (as shown in Table~\ref{tab:bert} and \cite{DBLP:journals/ipm/GuanZY24}). These findings reinforced our decision to adopt LLMs for this work.

It is worth noting that certain datasets used in our experiments are included in the pre-training data for Flan-T5. For detailed information, please refer to Table~\ref{tab:tab07}.

\subsection{Metrics}
In this work, we mainly consider performance at the sentence level. Therefore, we report average accuracy on the balanced boolean benchmarks \citep{brodersen2010balanced}, defined as:

\begin{equation}
    \begin{aligned}
        \text{Acc} = \frac{1}{|D|} \sum_{(x_i, y_i) \in D} F_C \left[ S(z(x_i)) = y_i \right]
    \end{aligned}
\end{equation}

\noindent where $|D|$ denotes the entire count of samples within the dataset, and $F_C$ represents the flag function which return a value of 1 when the condition enclosed in parentheses is satisfied, and 0 when it is not.
$z(x_i)$ is the logit output of the model for the input statement $x_i$ and $S(z(x_i)) = y_i$ is the sign function applied to the logit $z(x_i)$, and $y_i$ is the ground-truth label for the statement $x_i$, indicating whether the statement is actually True or False.

\subsection{Baseline Models}
We compared the CCSG with several state-of-the-art publicly accessible models that can be either directly utilized or adapted for commonsense statement verification. These models, presented in ascending order of performance, provide a comprehensive comparison. The baseline models include:

\vspace{2mm}

\noindent \textbf{SKD Critic.~~~} 
 It leverages the RoBERTa-large architecture \citep{liu2019roberta}, designed to discern erroneous commonsense knowledge produced by their SKD approach \citep{west2021symbolic}.

\vspace{2mm}

\noindent \textbf{I2D2 Critic.~~~} A critic model aimed at sieving out inaccurate commonsense knowledge that emerges from their I2D2 methodology \citep{bhagavatula2022i2d2}.

\vspace{2mm}

\noindent \textbf{UnifiedQA-v2.~~}
 A question-answering model and designed to handle datasets featuring diverse input formats, including those that are boolean in nature \citep{kadavath2022language}.

\vspace{2mm}

\noindent \textbf{Entailer.~~~} A model engineered to generate proof trees for hypotheses that are based on general scientific knowledge and understanding \citep{tafjord2022entailer}.

\vspace{2mm}
\noindent \textbf{VERA.~~~} The model includes two versions, VERA-T5 and VERA-LLama. In general, VERA-T5 outperforms VERA-LLama \citep{liu2023vera}.

\vspace{2mm}
\noindent \textbf{GPT3.5.~~~} A general-purpose autoregressive language models that are decoder-only\citep{openaib}. To purpose this model for the role of a commonsense verifier, the prompt follows previous work \citep{liu2023vera}.

\vspace{2mm}

\noindent \textbf{ChatGPT.} The successor to its predecessor GPT-3, exhibit substantial improvements in both the depth and breadth of their linguistic capabilities \citep{openaia, achiam2023gpt}.

\vspace{2mm}

\noindent \textbf{Flan-T5.} A collection of sequence-to-sequence language models that 
 include multiple encoder versions through instruction-based training.

\subsection{Hyperparameter Settings}
Our experimental work is carried out on a high-performance workstation with six NVIDIA L20 including 288G memory and the environment of torch 1.13.1. See the Table~\ref{tab:tab08} for more details of the training parameters. Furthermore, it is specifically noted that the integral gradient scores obtained in the preceding iteration are inaccessible during the first training epoch; hence, counterfactual samples constructor only computes the word contributions at the first training epoch.

\begin{table*}[!ht]
    \centering
    \footnotesize
    \renewcommand{\arraystretch}{1.1}
    \setlength{\tabcolsep}{1.2mm}{
    \caption{\label{tab:tab08}
        Hyperparameter settings.
    }
    \begin{tabular}{lll}
    \toprule
        \textbf{Symbol} & \textbf{Value} & \textbf{Description} \\ \hline
        L & 128 & Maximum token count per statement \\
        $B_G$ & 4 & Number of statements in each batch \\
        $B_S$ & 8 & Maximum number of statements allowed per batch \\
        $R_{p-drop}$ & 0.05 & Droptout generation ratio for positive samples \\
        $\eta_{T5}$ & $1 \times 10^{-5} $ & Learning rate for CCSG with T5 encoder backbone \\
        $\alpha$ & 1.0 & Weighting factor for binary classification loss \\
        $\beta$ & 0.25 & Weighting factor for contrastive loss \\
        $\tau$ & 0.05 & Temperature in contrastive loss \\ \bottomrule

    \end{tabular}}
\end{table*}

\section{Experimental Results}

In this part, we assess the capability of CCSG in determining the credibility of commonsense assertions and benchmark its performance against other baseline models. We showcase the effectiveness of CCSG across two separate contexts: distinguishing commonsense statements and filtering commonsense output produced by LLMs. We employ a threshold of $s = 0.5$ for predicting the accuracy of commonsense statements.

\begin{table*}[!ht]
    \centering
    \renewcommand{\arraystretch}{1}
    \setlength{\tabcolsep}{1.1mm}{
    \caption{\label{tab:tab01}
        Comparison result between our method with baselines in accuracy.
    }
    \begin{tabular}{lcccccccccc}
    \toprule
        \textbf{Dataset~~~~→} & \textbf{Avg} & OBQA & COPA & SciQ & QASC & SIQA & COHDA & ComVE & CSQA2 & C2S\\ \hline
        SKD Critic(355M) & 37.83   & 27.60  & 53.00  & 27.30  & 12.42  & 39.20  & 29.35  & 52.56  & 47.60  & 51.41 \\
        I2D2 Critic(355M) & 59.64   & 44.80  & 72.80  & 55.10  & 45.25  & 56.45  & 67.30  & 88.26  & 43.65  & 63.17 \\
        UnifiedQA-v2(11B) & 58.60   & 54.60  & 81.20  & 42.20  & 32.61  & 52.10  & 49.00  & 83.65  & 56.05  & 75.96 \\
        Entailer(11B) & 75.78   & 74.40  & 92.40  & 76.90  & 57.56  & 64.33  & 80.70  & 96.89  & 56.00  & 82.86   \\
        PPL(GPT-3.5) & 66.36   & 45.20  & / & 86.80  & 57.02  & 51.23  & / & 85.37  & / & 72.56   \\ 
        GPT-3.5(175B) & 77.74   & 74.20  & 87.00  & 86.00  & 62.85  & 65.30  & 85.05  & 97.39  & 60.55  & 81.33   \\
        ChatGPT & 62.63   & 60.80  & 58.80  & 60.70  & 42.01  & 52.20  & 56.75  & 93.08  & / & 76.73   \\ 
        ~~~~~~+5shot Cot & 67.51   & 62.40  & / & 69.77  & 47.52  & 52.25  & / & 90.98  & / & 82.14   \\ 
        GPT-4 & 71.89   & 76.00  & 64.00  & 70.00  & 44.00  & 57.00  & 66.00  & 95.00  & \textbf{81.00}  & \textbf{94.00}   \\
        ~~~~~~+5shot Cot & 75.80   & 79.80  & / & 80.00  & 44.00  & 67.00  & / & 92.08  & / & 91.92   \\
        Flan-T5(11B) & 80.65   & 79.60  & 93.00  & 80.80  & 64.58  & 73.23  & 89.60  & \textbf{98.40}  & 62.25  & 84.40   \\ 
        VERA+LLaMa(7B) & 83.08   & 80.20  & 91.80  & 90.00  & 71.38  & 79.89  & 88.95  & 97.99  & 63.85  & 83.63   \\
        VERA+T5(5B) & 84.42   & 83.20  & 93.40  & 88.80  & 73.33  & 80.14  & 88.60  & 97.79  & 68.60  & 85.93   \\ \hline
        CCSG+T5(5B) & \textbf{87.49} & \textbf{89.99} & \textbf{93.51} & \textbf{91.03} & \textbf{90.08} & \textbf{91.14} & \textbf{91.01} & 92.77 & 70.37 & 82.95  \\ \bottomrule
    \end{tabular}}
\end{table*}

\begin{table*}[!h]
    \centering
    \footnotesize
    \renewcommand{\arraystretch}{1.2}
    \setlength{\tabcolsep}{1.2mm}{
    \caption{\label{tab:alb}
        Results of ablation experiments.
    }
    \begin{tabular}{lcccccccccc}
    \toprule
        \textbf{Dataset} & \textbf{Avg} & OBQA & COPA & SciQ & QASC & SIQA & COHDA & ComVE & CSQA2 & C2S\\ \hline
        CCSG+T5(5B) & \textbf{87.49} & \textbf{89.99} & \textbf{93.51} & \textbf{91.03} & \textbf{90.08} & \textbf{91.14} & \textbf{91.01} & 88.86 & \textbf{70.37} & \textbf{81.43}  \\ \hline
        ~~~~~~\emph{w/o} CCSG & 81.11 & 81.52 & 85.21 & 80.01 & 70.05 & 78.72 & 86.53 & \textbf{90.76} & 60.10 & 79.33\\ \bottomrule
    \end{tabular}}
\end{table*}

\begin{table*}[!ht]
    \centering
    \renewcommand{\arraystretch}{1.2}
    \setlength{\tabcolsep}{1.2mm}{
    \caption{\label{tab:tab02}
        Accuracy results of filtering commonsense output produced by LLMs.
    }
    \begin{tabular}{lllcccccc}
    \toprule
        Generator & Filter & QA & \textbf{Avg} & CSQA & QASC & PIQA & WG & OBQA \\ \hline
        \# & \# & UnifiedQA-large & 58.35   & 61.43  & 43.09  & 63.66  & 53.35  & 70.20   \\ 
        GPT-3(davinci) & \# & UnifiedQA-large & 66.59   & 70.19  & 63.82  & 67.74  & 56.59  & 74.60   \\ 
        GPT-3(davinci) & VERA & UnifiedQA-large & 68.66   & 71.91  & 66.20  & \textbf{70.35} & 57.22  & 77.60   \\ \hline
        GPT-3(davinci) & Our & UnifiedQA-large & \textbf{70.60} & \textbf{77.32} & \textbf{67.8}6 & 68.95 & \textbf{58.90} & \textbf{79.99} \\ \bottomrule
    \end{tabular}}
\end{table*}

\subsection{Distinguishing Commonsense Statements}

The plausibility scores from CCSG can apply to multiple-choice and boolean commonsense tasks. Initially, we transform these tasks into the statement group format as described by \citep{liu2023vera}. Table~\ref{tab:tab01} presents the outcomes when utilizing CCSG for resolving commonsense challenges. Generally, we have the following findings:

CCSG outperforms previous baselines by large margins, including previous state-of-the-art \citep{liu2023vera}, VERA+T5, by 3.07$\%$ on (absolute) average accuracy. In addition, CCSG beats GPT-4 (+5shot Cot) by 11.69$\%$ on average accuracy. However, it is noted that CCSG exhibits suboptimal performance across the ComVE, CSQA2, and C2S datasets. This underperformance may stem from the interference of negative samples generated by Contrastive Cross-Sample Generation (CCSG) in low sample size or domain-specific scenarios. Generated samples, intended to enhance model discrimination, can introduce Confounding noise that disrupts training, particularly when the sample size is small or the domain is highly specialized. Furthermore, we have conducted ablation studies to underscore the efficacy of CCSG. Utilizing CCSG, our refined model exhibits a significant enhancement of 6.38$\%$ in terms of Average Precision when contrasted with the original T5 model. The findings from the ablation study are presented in Table~\ref{tab:alb}.

Generally, when dealing with boolean datasets for binary classification, it is crucial to choose a suitable decision threshold. However, our analysis reveals that a logit value of zero $(z = 0)$ often aligns closely with the optimal threshold for distinguishing between accurate and inaccurate commonsense assertions. Consequently, we forego the estimation of a model-specific threshold and opt to use a threshold of $z = 0$.

\subsection{Filtering LLM-generated Commonsense Knowledge}
 We find that the application of CCSG for filtering commonsense knowledge significantly enhances the efficacy of knowledge-augmented reasoning approaches. Within the GKP framework proposed by \citep{liu2021generated}, the resolution of commonsense QA tasks is structured as a two-step process: initially, a knowledge model synthesizes a set of pertinent commonsense knowledge statements in response to the query at hand; subsequently, a dedicated QA model formulates its predictions by leveraging this generated knowledge. A significant challenge that undermines the efficacy of the Generated Knowledge Prompting framework is the potential for the knowledge model to produce non-factual statements. The incorporation of inaccurate knowledge can misguide the subsequent QA model. To address this issue, \citep{liu2023vera} suggest incorporating VERA as a filter for the knowledge statements generated before they are employed by the LLM-QA system. Specifically, they retain only those statements that their model assigns a plausibility score exceeding 0.5, ensuring a higher degree of factual accuracy in the knowledge base provided to the LLM-QA model.

 Following the previous research \citep{liu2022rainier}, we employ UnifiedQA-large as the QA model and treat few-shot GPT-3 (davinci) \citep{brown2020language} as the knowledge model. We follow the evaluation settings in \citep{liu2022rainier}. Results are presented in Table~\ref{tab:tab02}. 
 Incorporating CCSG for knowledge filtering significantly boosts the mean accuracy of GPT-3's knowledge by 4.01$\%$, which is a 1.94$\%$ improvement over VERA. CCSG demonstrates remarkable efficacy in overseeing and enhancing the quality of commonsense knowledge produced by the expansive GPT-3 (davinci) model.

\subsection{Validation of Effectiveness}
For the motivations behind CCSG, we conduct detailed analyses to further reveal why CCSG works and how it enhances model performance, ultimately aiming to uncover its core advantages and implications for PE model.

\subsubsection{Alleviating Commonsense Bias}
To substantiate that CCSG can alleviate commonsense bias with
the generated counterfactual samples, we randomly select 150 sentences in four fields with mixed commonsense errors and evaluated them by CCSG. After that, we manually checked the commonsense errors and calculated the deviation rate. The results presented in Table~\ref{tab:tab3} indicate that CCSG exhibits a much lower bias rate than the vanilla T5-only-encoder model. Thus, CCSG can significantly enhance commonsense-sensitive and mitigate the commonsense bias.

\begin{figure}
    \centering
    \includegraphics[width=0.5\linewidth]{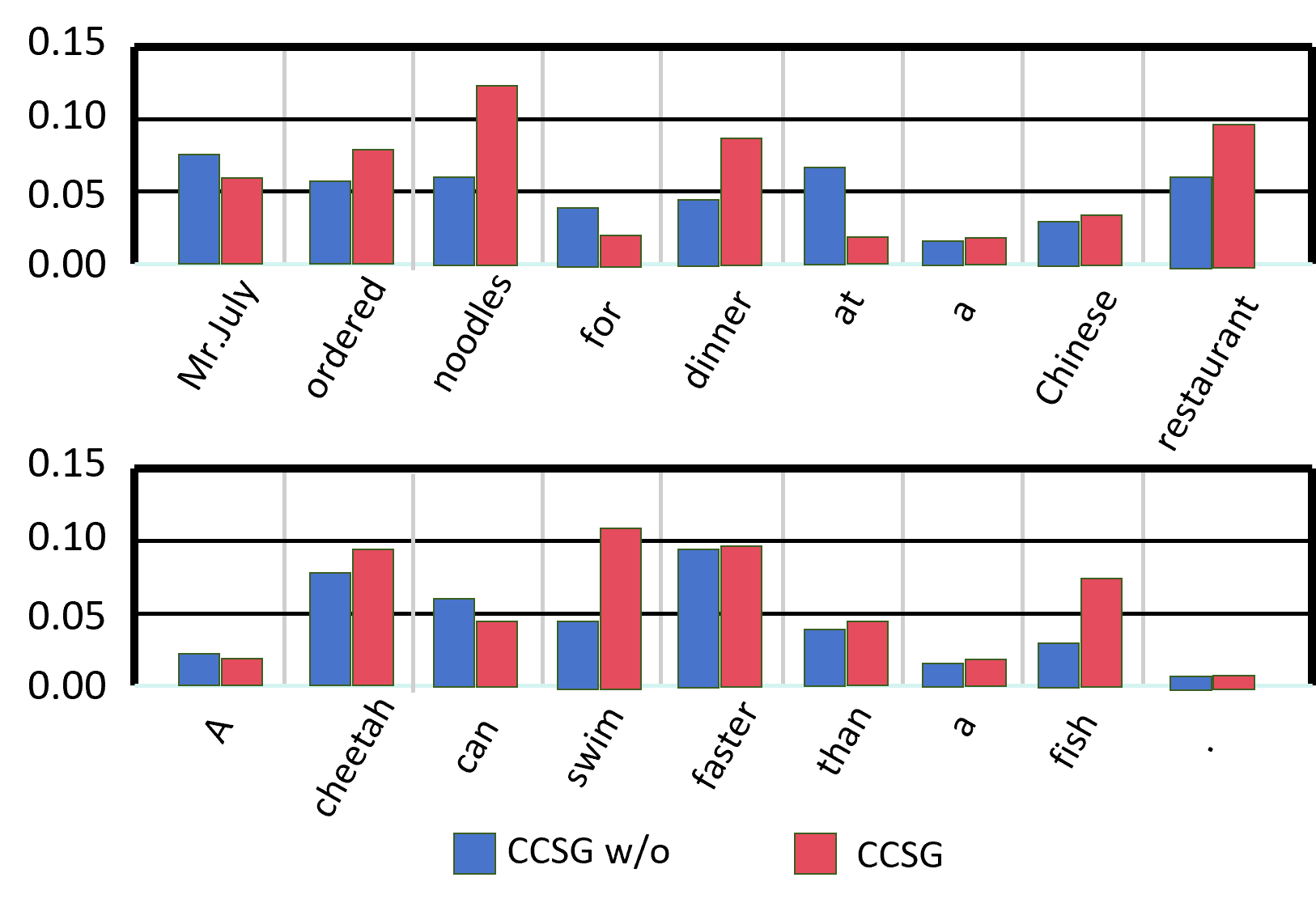}
    \caption{Visualization of the contribution distribution of each token.}
    \label{fig:fig04}
\end{figure}

\begin{table}[!ht]
    \centering
    \renewcommand{\arraystretch}{1.2}
    \setlength{\tabcolsep}{1.2mm}{
    \caption{\label{tab:tab3}
        The commonsense bias rates issued by Vanilla T5 (\emph{w/o} CCSG) and CCSG in four specific fields.
    }
    \begin{tabular}{ccc}
    \toprule
       \textbf{~~Field~~} & \textbf{Vanilla~T5~~} & \textbf{CCSG~~} \\ \hline
       Physics & 40$\%$ & 24$\%$ \\
       Food & 21$\%$ & 4$\%$ \\
       Math & 47$\%$ & 36$\%$ \\
       Time Series & 25$\%$ & 9$\%$ \\ \bottomrule
    \end{tabular}}
\end{table}

\subsubsection{Enhancing Language Explainability}
To explore the rationale behind the counterfactual constructor, we concentrate on the distribution of contribution for each word, as it indicates the importance of each word in affecting the prediction. Following previous work \citep{chen2023counterfactual}, we calculate the contribution distribution of each word (see Fig. \ref{fig:fig04}). We discover that the words of  ``\emph{Mr.July}'' and ``\emph{at}'' have higher contribution in the original model. However, after adopting our CCSG, the contribution scores of ``\emph{noodles}'', ``\emph{dinner}'' and ``\emph{restaurant}'' are upper than before, and thus can mildly help the model to strengthen the language-explainable.

\section{Conclusion and Limitation}

In this research, we introduced the CCSG method, a novel and model-agnostic approach designed to enhance the language-explainable and commonsense-sensitive abilities of PE models. By leveraging counterfactual reasoning and contrastive learning, CCSG effectively mitigates the over-reliance on superficial linguistic correlations often observed in existing PE models. Specifically, our method constructs positive and negative counterfactual samples through word-level contribution analysis and dropout-based augmentation, guiding language models to focus on relevant linguistic regions while improving sensitivity to commonsense variations. The proposed CCSG framework also incorporates SCMs to address commonsense biases from a causal perspective, serving as a versatile mediator for improving PE tasks. Extensive experiments on nine datasets demonstrated that CCSG not only reduces spurious correlations but also achieves a 3.07\% improvement in plausibility estimation performance, surpassing existing benchmarks. These results highlight its potential to enhance robustness, interpretability, and fairness in PE tasks.

Despite its effectiveness, CCSG has certain limitations. It is designed to predict the plausibility of statements based on existing commonsense knowledge of the real world and struggles with predicting statements that extend beyond reality (e.g., fictional or fantastical contexts). Furthermore, CCSG lacks moral discernment, and toxic inputs may compromise the accuracy and reliability of its predictions. It is essential to clarify that CCSG outputs do not represent the opinions or viewpoints of its authors. As a research-oriented prototype, CCSG is intended primarily for academic exploration and is not suitable for direct deployment in real-world decision-making scenarios with significant consequences. Future work will focus on addressing these limitations by exploring extensions to fictional contexts, incorporating mechanisms for ethical reasoning, and improving scalability for broader applications.

\printcredits

\section*{Acknowledgment}
This work presents the results of a research project partially funded by the China Fundamental Research Funds for the Central Universities (Grant Nos. 2662022XXYJ001 and 2662023XXPY005). We also acknowledge the support of the Australian Research Council under Grant DP230101122.

\bibliographystyle{apalike}
\bibliography{cas-refs}

\begin{thebibliography}{}

\bibitem[Achiam et~al., 2023]{achiam2023gpt}
Achiam, J., Adler, S., Agarwal, S., Ahmad, L., Akkaya, I., Aleman, F.~L.,
  Almeida, D., Altenschmidt, J., Altman, S., Anadkat, S., et~al. (2023).
\newblock Gpt-4 technical report.
\newblock {\em arXiv preprint arXiv:2303.08774}.

\bibitem[Bender and Koller, 2020]{bender2020climbing}
Bender, E.~M. and Koller, A. (2020).
\newblock Climbing towards nlu: On meaning, form, and understanding in the age
  of data.
\newblock In {\em Proceedings of the 58th annual meeting of the association for
  computational linguistics}, pages 5185--5198.

\bibitem[Bhagavatula et~al., 2022]{bhagavatula2022i2d2}
Bhagavatula, C., Hwang, J.~D., Downey, D., Bras, R.~L., Lu, X., Qin, L.,
  Sakaguchi, K., Swayamdipta, S., West, P., and Choi, Y. (2022).
\newblock I2d2: Inductive knowledge distillation with neurologic and
  self-imitation.
\newblock {\em arXiv preprint arXiv:2212.09246}.

\bibitem[Brodersen et~al., 2010]{brodersen2010balanced}
Brodersen, K.~H., Ong, C.~S., Stephan, K.~E., and Buhmann, J.~M. (2010).
\newblock The balanced accuracy and its posterior distribution.
\newblock In {\em 2010 20th international conference on pattern recognition},
  pages 3121--3124. IEEE.

\bibitem[Brown, 2020]{brown2020language}
Brown, T.~B. (2020).
\newblock Language models are few-shot learners.
\newblock {\em arXiv preprint arXiv:2005.14165}.

\bibitem[Chen et~al., 2021]{chen2021explaining}
Chen, H., Feng, S., Ganhotra, J., Wan, H., Gunasekara, C., Joshi, S., and Ji,
  Y. (2021).
\newblock Explaining neural network predictions on sentence pairs via learning
  word-group masks.
\newblock {\em arXiv preprint arXiv:2104.04488}.

\bibitem[Chen et~al., 2023a]{chen2023counterfactual}
Chen, L., Zheng, Y., Niu, Y., Zhang, H., and Xiao, J. (2023a).
\newblock Counterfactual samples synthesizing and training for robust visual
  question answering.
\newblock {\em IEEE Transactions on Pattern Analysis and Machine Intelligence},
  45(11):13218--13234.

\bibitem[Chen et~al., 2023b]{chen2023causal}
Chen, Z., Hu, L., Li, W., Shao, Y., and Nie, L. (2023b).
\newblock Causal intervention and counterfactual reasoning for multi-modal fake
  news detection.
\newblock In {\em Proceedings of the 61st Annual Meeting of the Association for
  Computational Linguistics (Volume 1: Long Papers)}, pages 627--638.

\bibitem[Cheng et~al., 2024]{cheng2024data}
Cheng, D., Li, J., Liu, L., Liu, J., and Le, T.~D. (2024).
\newblock Data-driven causal effect estimation based on graphical causal
  modelling: A survey.
\newblock {\em ACM Computing Surveys}, 56(5):1--37.

\bibitem[Cheng et~al., 2023]{cheng2023causal}
Cheng, D., Xu, Z., Li, J., Liu, L., Liu, J., and Le, T.~D. (2023).
\newblock Causal inference with conditional instruments using deep generative
  models.
\newblock In {\em Proceedings of the AAAI conference on artificial
  intelligence}, volume~37, pages 7122--7130.

\bibitem[Eisenstein, 2022]{eisenstein2022informativeness}
Eisenstein, J. (2022).
\newblock Informativeness and invariance: Two perspectives on spurious
  correlations in natural language.
\newblock {\em arXiv preprint arXiv:2204.04487}.

\bibitem[Feng et~al., 2023]{feng2023less}
Feng, T., Qu, L., and Haffari, G. (2023).
\newblock Less is more: Mitigate spurious correlations for open-domain dialogue
  response generation models by causal discovery.
\newblock {\em Transactions of the Association for Computational Linguistics},
  11:511--530.

\bibitem[Gao et~al., 2021]{gao2021simcse}
Gao, T., Yao, X., and Chen, D. (2021).
\newblock Simcse: Simple contrastive learning of sentence embeddings.
\newblock {\em arXiv preprint arXiv:2104.08821}.

\bibitem[Goyal et~al., 2020]{goyal2020cam}
Goyal, N., Paneri, R., Agarwal, A., Kalani, U., Sancheti, A., and Chhaya, N.
  (2020).
\newblock Cam-gen: Causally-aware metric-guided text generation.
\newblock {\em arXiv preprint arXiv:2010.12795}.

\bibitem[Guan et~al., 2024]{DBLP:journals/ipm/GuanZY24}
Guan, B., Zhu, X., and Yuan, S. (2024).
\newblock A t5-based interpretable reading comprehension model with more
  accurate evidence training.
\newblock {\em Inf. Process. Manag.}, 61(2):103584.

\bibitem[Honnibal and Montani, 2017]{honnibal2017spacy}
Honnibal, M. and Montani, I. (2017).
\newblock spacy 2: Natural language understanding with bloom embeddings,
  convolutional neural networks and incremental parsing.
\newblock {\em To appear}, 7(1):411--420.

\bibitem[Hu and Li, 2021]{hu2021causal}
Hu, Z. and Li, L.~E. (2021).
\newblock A causal lens for controllable text generation.
\newblock {\em Advances in Neural Information Processing Systems},
  34:24941--24955.

\bibitem[Jain and Wallace, 2019]{jain2019attention}
Jain, S. and Wallace, B.~C. (2019).
\newblock Attention is not explanation.
\newblock {\em arXiv preprint arXiv:1902.10186}.

\bibitem[Jia et~al., 2019]{jia2019certified}
Jia, R., Raghunathan, A., G{\"o}ksel, K., and Liang, P. (2019).
\newblock Certified robustness to adversarial word substitutions.
\newblock {\em arXiv preprint arXiv:1909.00986}.

\bibitem[Jung et~al., 2022]{jung2022maieutic}
Jung, J., Qin, L., Welleck, S., Brahman, F., Bhagavatula, C., Bras, R.~L., and
  Choi, Y. (2022).
\newblock Maieutic prompting: Logically consistent reasoning with recursive
  explanations.
\newblock {\em arXiv preprint arXiv:2205.11822}.

\bibitem[Kadavath et~al., 2022]{kadavath2022language}
Kadavath, S., Conerly, T., Askell, et~al. (2022).
\newblock Language models (mostly) know what they know.
\newblock {\em arXiv preprint arXiv:2207.05221}.

\bibitem[Keith et~al., 2020]{keith2020text}
Keith, K.~A., Jensen, D., and O'Connor, B. (2020).
\newblock Text and causal inference: A review of using text to remove
  confounding from causal estimates.
\newblock {\em arXiv preprint arXiv:2005.00649}.

\bibitem[Khosla et~al., 2020]{khosla2020supervised}
Khosla, P., Teterwak, P., Wang, C., Sarna, A., Tian, Y., Isola, P., Maschinot,
  A., Liu, C., and Krishnan, D. (2020).
\newblock Supervised contrastive learning.
\newblock {\em Advances in neural information processing systems},
  33:18661--18673.

\bibitem[Kiritchenko and Mohammad, 2018]{kiritchenko2018examining}
Kiritchenko, S. and Mohammad, S.~M. (2018).
\newblock Examining gender and race bias in two hundred sentiment analysis
  systems.
\newblock {\em arXiv preprint arXiv:1805.04508}.

\bibitem[Ling et~al., 2023]{ling2023knowledge}
Ling, C., Zhang, X., Zhao, X., Wu, Y., Liu, Y., Cheng, W., Chen, H., and Zhao,
  L. (2023).
\newblock Knowledge-enhanced prompt for open-domain commonsense reasoning.
\newblock In {\em 1st AAAI Workshop on Uncertainty Reasoning and Quantification
  in Decision Making}.

\bibitem[Liu et~al., 2022]{liu2022rainier}
Liu, J., Hallinan, S., Lu, X., He, P., Welleck, S., Hajishirzi, H., and Choi,
  Y. (2022).
\newblock Rainier: Reinforced knowledge introspector for commonsense question
  answering.
\newblock {\em arXiv preprint arXiv:2210.03078}.

\bibitem[Liu et~al., 2021]{liu2021generated}
Liu, J., Liu, A., Lu, X., Welleck, S., West, P., Bras, R.~L., Choi, Y., and
  Hajishirzi, H. (2021).
\newblock Generated knowledge prompting for commonsense reasoning.
\newblock {\em arXiv preprint arXiv:2110.08387}.

\bibitem[Liu et~al., 2023]{liu2023vera}
Liu, J., Wang, W., Wang, D., Smith, N.~A., Choi, Y., and Hajishirzi, H. (2023).
\newblock Vera: A general-purpose plausibility estimation model for commonsense
  statements.
\newblock {\em arXiv preprint arXiv:2305.03695}.

\bibitem[Liu, 2019]{liu2019roberta}
Liu, Y. (2019).
\newblock Roberta: A robustly optimized bert pretraining approach.
\newblock {\em arXiv preprint arXiv:1907.11692}.

\bibitem[Madaan et~al., 2021]{madaan2021generate}
Madaan, N., Padhi, I., Panwar, N., and Saha, D. (2021).
\newblock Generate your counterfactuals: Towards controlled counterfactual
  generation for text.
\newblock In {\em Proceedings of the AAAI Conference on Artificial
  Intelligence}, volume~35, pages 13516--13524.

\bibitem[Marcus and Davis, 2023]{marcus_davis_2023}
Marcus, G. and Davis, E. (2023).
\newblock Chatgpt/llm errors.
\newblock (public).

\bibitem[Mu and Li, 2023]{mu2023enhancing}
Mu, F. and Li, W. (2023).
\newblock Enhancing event causality identification with counterfactual
  reasoning.
\newblock In {\em Proceedings of the 61st Annual Meeting of the Association for
  Computational Linguistics (Volume 2: Short Papers)}, pages 967--975.

\bibitem[OpenAI, 2022a]{openaia}
OpenAI (2022a).
\newblock Introducing chatgpt.
\newblock (public).

\bibitem[OpenAI, 2022b]{openaib}
OpenAI (2022b).
\newblock Moddels - overview - gpt3.5.
\newblock (public).

\bibitem[Pearl, 2009a]{pearl2009causal}
Pearl, J. (2009a).
\newblock Causal inference in statistics: An overview.

\bibitem[Pearl, 2009b]{pearl2009causality}
Pearl, J. (2009b).
\newblock {\em Causality}.
\newblock Cambridge university press.

\bibitem[Pearl et~al., 2000]{pearl2000models}
Pearl, J. et~al. (2000).
\newblock Models, reasoning and inference.
\newblock {\em Cambridge, UK: CambridgeUniversityPress}, 19(2):3.

\bibitem[Pennington et~al., 2014]{pennington2014glove}
Pennington, J., Socher, R., and Manning, C.~D. (2014).
\newblock Glove: Global vectors for word representation.
\newblock In {\em Proceedings of the 2014 conference on empirical methods in
  natural language processing (EMNLP)}, pages 1532--1543.

\bibitem[Raffel et~al., 2020]{raffel2020exploring}
Raffel, C., Shazeer, N., Roberts, A., Lee, K., Narang, S., Matena, M., Zhou,
  Y., Li, W., and Liu, P.~J. (2020).
\newblock Exploring the limits of transfer learning with a unified text-to-text
  transformer.
\newblock {\em Journal of machine learning research}, 21(140):1--67.

\bibitem[Ribeiro et~al., 2020]{ribeiro2020beyond}
Ribeiro, M.~T., Wu, T., Guestrin, C., and Singh, S. (2020).
\newblock Beyond accuracy: Behavioral testing of nlp models with checklist.
\newblock {\em arXiv preprint arXiv:2005.04118}.

\bibitem[Roberts et~al., 2020]{roberts2020adjusting}
Roberts, M.~E., Stewart, B.~M., and Nielsen, R.~A. (2020).
\newblock Adjusting for confounding with text matching.
\newblock {\em American Journal of Political Science}, 64(4):887--903.

\bibitem[Roemmele et~al., 2011]{roemmele2011choice}
Roemmele, M., Bejan, C.~A., and Gordon, A.~S. (2011).
\newblock Choice of plausible alternatives: An evaluation of commonsense causal
  reasoning.
\newblock In {\em 2011 AAAI spring symposium series}.

\bibitem[Ross et~al., 2017]{ross2017right}
Ross, A.~S., Hughes, M.~C., and Doshi-Velez, F. (2017).
\newblock Right for the right reasons: Training differentiable models by
  constraining their explanations.
\newblock {\em arXiv preprint arXiv:1703.03717}.

\bibitem[Selvaraju et~al., 2020]{selvaraju2020grad}
Selvaraju, R.~R., Cogswell, M., Das, A., Vedantam, R., Parikh, D., and Batra,
  D. (2020).
\newblock Grad-cam: visual explanations from deep networks via gradient-based
  localization.
\newblock {\em International journal of computer vision}, 128:336--359.

\bibitem[Selvaraju et~al., 2019]{selvaraju2019taking}
Selvaraju, R.~R., Lee, S., Shen, Y., Jin, H., Ghosh, S., Heck, L., Batra, D.,
  and Parikh, D. (2019).
\newblock Taking a hint: Leveraging explanations to make vision and language
  models more grounded.
\newblock In {\em Proceedings of the IEEE/CVF international conference on
  computer vision}, pages 2591--2600.

\bibitem[Tafjord et~al., 2022]{tafjord2022entailer}
Tafjord, O., Mishra, B.~D., and Clark, P. (2022).
\newblock Entailer: Answering questions with faithful and truthful chains of
  reasoning.
\newblock {\em arXiv preprint arXiv:2210.12217}.

\bibitem[Talmor et~al., 2018]{talmor2018commonsenseqa}
Talmor, A., Herzig, J., Lourie, N., and Berant, J. (2018).
\newblock Commonsenseqa: A question answering challenge targeting commonsense
  knowledge.
\newblock {\em arXiv preprint arXiv:1811.00937}.

\bibitem[Thorne et~al., 2018]{thorne2018fever}
Thorne, J., Vlachos, A., Christodoulopoulos, C., and Mittal, A. (2018).
\newblock Fever: a large-scale dataset for fact extraction and verification.
\newblock {\em arXiv preprint arXiv:1803.05355}.

\bibitem[Tokpo and Calders, 2024]{tokpo2024fairflow}
Tokpo, E.~K. and Calders, T. (2024).
\newblock Fairflow: An automated approach to model-based counterfactual data
  augmentation for nlp.
\newblock In {\em Joint European Conference on Machine Learning and Knowledge
  Discovery in Databases}, pages 160--176. Springer.

\bibitem[Udomcharoenchaikit et~al., 2022]{udomcharoenchaikit2022mitigating}
Udomcharoenchaikit, C., Ponwitayarat, W., Payoungkhamdee, P., Masuk, K.,
  Buaphet, W., Chuangsuwanich, E., and Nutanong, S. (2022).
\newblock Mitigating spurious correlation in natural language understanding
  with counterfactual inference.
\newblock In {\em Proceedings of the 2022 Conference on Empirical Methods in
  Natural Language Processing}, pages 11308--11321.

\bibitem[Wang et~al., 2025]{DBLP:journals/ipm/WangSMXWYS25}
Wang, A., Song, L., Min, Z., Xu, G., Wang, X., Yao, J., and Su, J. (2025).
\newblock Mitigating the negative impact of over-association for conversational
  query production.
\newblock {\em Inf. Process. Manag.}, 62(1):103907.

\bibitem[Wang et~al., 2023]{wang2023causal}
Wang, F., Mo, W., Wang, Y., Zhou, W., and Chen, M. (2023).
\newblock A causal view of entity bias in (large) language models.
\newblock {\em arXiv preprint arXiv:2305.14695}.

\bibitem[Wang and Culotta, 2020]{wang2020identifying}
Wang, Z. and Culotta, A. (2020).
\newblock Identifying spurious correlations for robust text classification.
\newblock {\em arXiv preprint arXiv:2010.02458}.

\bibitem[Wang and Culotta, 2021]{wang2021robustness}
Wang, Z. and Culotta, A. (2021).
\newblock Robustness to spurious correlations in text classification via
  automatically generated counterfactuals.
\newblock In {\em Proceedings of the AAAI Conference on Artificial
  Intelligence}, volume~35, pages 14024--14031.

\bibitem[West et~al., 2021]{west2021symbolic}
West, P., Bhagavatula, C., Hessel, J., Hwang, J.~D., Jiang, L., Bras, R.~L.,
  Lu, X., Welleck, S., and Choi, Y. (2021).
\newblock Symbolic knowledge distillation: from general language models to
  commonsense models.
\newblock {\em arXiv preprint arXiv:2110.07178}.

\bibitem[Wood-Doughty et~al., 2018]{wood2018challenges}
Wood-Doughty, Z., Shpitser, I., and Dredze, M. (2018).
\newblock Challenges of using text classifiers for causal inference.
\newblock In {\em Proceedings of the Conference on Empirical Methods in Natural
  Language Processing. Conference on Empirical Methods in Natural Language
  Processing}, volume 2018, page 4586. NIH Public Access.

\bibitem[Ze{\v{c}}evi{\'c} et~al., 2023]{zevcevic2023causal}
Ze{\v{c}}evi{\'c}, M., Willig, M., Dhami, D.~S., and Kersting, K. (2023).
\newblock Causal parrots: Large language models may talk causality but are not
  causal.
\newblock {\em arXiv preprint arXiv:2308.13067}.

\bibitem[Zeng et~al., 2020]{zeng2020counterfactual}
Zeng, X., Li, Y., Zhai, Y., and Zhang, Y. (2020).
\newblock Counterfactual generator: A weakly-supervised method for named entity
  recognition.
\newblock In {\em Proceedings of the 2020 Conference on Empirical Methods in
  Natural Language Processing (EMNLP)}, pages 7270--7280.

\bibitem[Zhao et~al., 2024]{DBLP:journals/ipm/ZhaoXJT24}
Zhao, Y., Xia, T., Jiang, Y., and Tian, Y. (2024).
\newblock Enhancing inter-sentence attention for semantic textual similarity.
\newblock {\em Inf. Process. Manag.}, 61(1):103535.

\end{thebibliography}

\section*{Declaration of competing interest}
The authors declare that they have no known competing financial interests or personal relationships that could have appeared to influence the work reported in this paper.

\end{document}